\documentclass[journal]{IEEEtran}
\ifCLASSINFOpdf
  \usepackage[pdftex]{graphicx}
  \graphicspath{{bibtex/}}
  \DeclareGraphicsExtensions{.pdf,.jpeg,.png}
\else
\fi
\usepackage{amsmath}
\usepackage{amssymb}

\usepackage{algorithmic}

\usepackage{array}
\usepackage{url}

\usepackage{enumitem}

\usepackage{xcolor}
\usepackage{hyperref}

\DeclareMathOperator*{\argmax}{argmax}

\begin{document}
\title{A Hybrid Deep Learning Model for Robust Biometric Authentication from Low-Frame-Rate PPG Signals}

\author{Arfina Rahman,~\IEEEmembership{Student Member,~IEEE,}
        and~Mahesh Banavar,~\IEEEmembership{Senior Member,~IEEE,}

\thanks{Manuscript received Month Day, Year; revised Month Day, Year. 
        (Corresponding author: First Author.)}
\thanks{The authors are with the Department of Electrical and Computer
Engineering, Clarkson University, Potsdam, NY 13676 USA (e-mail:
arahma@clarkson.edu; mbanavar@clarkson.edu).}

}

\maketitle

\begin{abstract}
Photoplethysmography (PPG) signals which measure changes in blood volume in the skin using light have recently gained attention in biometric authentication because of their non-invasive acquisition, inherent liveness detection, and suitability for low-cost wearable devices. However, PPG signals reliability is challenged by motion artifacts, illumination changes, and inter-subject physiological variability, making robust feature extraction and classification crucial. This study proposes a lightweight and cost-effective biometric authentication framework based on PPG signals extracted from low-frame-rate fingertip videos. The CFISHR dataset, comprising PPG recordings from 46 subjects at a sampling rate of 14 Hz, is employed for evaluation. The raw PPG signals undergo a structured preprocessing pipeline involving baseline drift removal, motion artifact suppression using Principal Component Analysis (PCA), bandpass filtering, Fourier-based resampling, and amplitude normalization. To generate robust representations, each one-dimensional PPG segment is converted into a two-dimensional time–frequency scalogram via the Continuous Wavelet Transform (CWT), effectively capturing transient cardiovascular dynamics. We developed a hybrid deep learning model, termed CVT–ConvMixer–LSTM, by combining spatial features from the Convolutional Vision Transformer (CVT) and ConvMixer branches with temporal features from a Long Short-Term Memory network (LSTM). The experimental results on 46 subjects demonstrate an authentication accuracy of  98\%, validating the robustness of the model to noise and variability between subjects. Due to its efficiency, scalability, and inherent liveness detection capability, the proposed system is well-suited for real-world mobile and embedded biometric security applications.
\end{abstract}

\begin{IEEEkeywords}
attention mechanisms; biometric authentication; feature extraction; secure identification; wavelet transform. 
\end{IEEEkeywords}

%
\IEEEpeerreviewmaketitle

\section{Introduction}
%
%
%
%
\IEEEPARstart{B}{iometric} authentication leverages unique physiological or behavioral characteristics to validate individual identity, enhancing the security and usability of digital systems~\cite{lienChallengesOpportunitiesBiometric2023}. Traditional modalities such as fingerprints, facial features, iris, retina, and voice have long been utilized due to their distinctiveness and reliability across various platforms~\cite{saiaInfluencingBrainWaves2023}. Recent advancements in sensing technologies and machine learning have enabled the exploration of alternative biometric signals like photoplethysmography (PPG) \cite{hnoohomPhysicalActivityRecognition2023}. PPG is a non-invasive optical technique that captures pulsatile blood volume changes in vascular tissue using light-based sensors. Acquired from sites like the fingertip or wrist, PPG signals encode cardiovascular dynamics that are inherently influenced by individual physiological traits. These individualized hemodynamic patterns position PPG as a viable and discriminative biometric signal \cite{armanac-julianReliabilityPulsePhotoplethysmography2022}.

\begin{figure}[t]
    \centering
    \includegraphics[width=0.95\columnwidth]{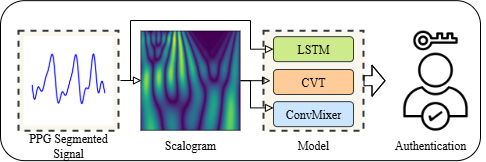}
    \caption{Authentication System Overview}
    \label{fig:auth_sys}
\end{figure}

\subsection{Literature Review}
Recent advancements in deep learning have significantly enhanced the performance and applicability of PPG based biometric authentication. The integration of convolutional and recurrent architectures enables the extraction of complex temporal and morphological representations for continuous and robust identity verification. Recent developments \cite{qureshiMedicalImageSegmentation2023, durmazOnDeviceDeepLearning2023} in deep learning-based biometric authentication have emphasized optimizing existing models like Convolutional Neural Networks (CNNs) and Recurrent Neural Networks (RNNs)~\cite{abbasClassificationPostCovid2023} for resource-constrained mobile and wearable environments, focusing on performance and real-time deployment. In~\cite{hazratifardEnsembleSiameseNetwork2023}, an Ensemble Siamese Network (ESN) was introduced for Electrocardiogram (ECG)-based continuous authentication in healthcare IoT (Internet of Things), addressing challenges such as data imbalance and new user enrollment. The system achieved strong performance, with accuracies of 93.6\% on the ECG-ID dataset and 96.8\% on the PTB (Physikalisch-Technische Bundesanstalt) dataset, and corresponding equal error rates (EERs) of 1.76\% and 1.69\%, respectively. Recent studies have leveraged CNN-based deep learning approaches for PPG signal analysis, focusing on robust biometric authentication and heart rate estimation in real-world conditions. The PPG-DaLiA dataset~\cite{reissDeepPPGLargeScale2019} supports this with large-scale synchronized data, while enhancements like variation-stable methods~\cite{hwangVariationStableFusionPPGBased2021}, end-to-end raw PPG processing~\cite{luqueENDtoENDPhotoplethYsmographY2018}, and rigorous training techniques improve performance under motion artifacts and cross-session variability. Reported authentication accuracy across studies ranges from approximately 78\% to over 98\%, depending on session type and dataset. Despite promising results, challenges remain in generalizing across subjects and conditions~\cite{siamBiosignalClassification2021}, as most prior work relies on signals collected under controlled settings with high-quality sensors. 

Some studies explore privacy-preserving and time-stable PPG-based biometric authentication.  Homomorphic Random Forest (HRF) was used in~\cite{zhangSecureFlexiblePPGBased2024}, to classify encrypted PPG features, achieving 96.4\% accuracy and 2.14\% EER across five datasets while ensuring data privacy and robustness across different sensor modalities. Similarly,~\cite{hwangEvaluationTimeStability2021} proposed a CNN-LSTM hybrid model incorporating time-stretching techniques for temporal stability, yielding 98\% accuracy in single sessions and 87.1\% across sessions, though at the cost of higher inference complexity. The Bi-LSTM approach in~\cite{Ortiz2414-4088}, demonstrated 96.7\% accuracy on smartphone PPG, underscoring the importance of temporal modeling. A Siamese 1D network in \cite{seokPhotoplethysmogramBiometricAuthentication2023} leveraged multicycle averaging for noise suppression, achieving an accuracy of 97.23\%  and an AUC (Area Under the Curve) of 0.98, while \cite{liuDualdomainMultiscaleFusion2023} proposed a dual-domain multiscale fusion network (DMFDNN) combining temporal and spectral features, yielding 96\% accuracy and showing strong generalization across four datasets. An XGBoost classifier using handcrafted features extraction in \cite{ortizTimeSeriesForecastingExtreme2022} attained 96.38\% accuracy but highlighted trade-offs in F1 score (72\%) and precision (67\%), suggesting room for better class balance. While these methods achieved better results compared to traditional handcrafted approaches, they often remain constrained by signal instability, incomplete feature utilization, or computational overhead. 

Building upon these trends in privacy-preserving and time-stable PPG-based biometric authentication, several recent studies have integrated federated learning and multi-modal frameworks to enhance security, scalability, and practicality in biometric authentication. A federated dual CNN framework ~\cite{coelhoMultimodalBiometricAuthentication2023}, enabled secure and decentralized PPG–ECG verification without sharing raw data, supporting privacy-preserving collaboration across devices. Similarly,~\cite{puNovelRobustPhotoplethysmogramBased2022}, implemented a real-time PPG-based authentication system, achieving 98\% identification accuracy and a 5.5\% EER, validating on-device feasibility. A multimodal model in~\cite{AhamedFutureInternet} further demonstrated 99.8\% accuracy and 0.16\% EER by combining ECG and PPG signals, underscoring the promise of multimodal biometric frameworks. However, high-performing frameworks such as CNN–LSTM, Bi-LSTM, and multimodal models, often remain computationally intensive, while privacy-preserving models further increase complexity. Thus, developing a lightweight, generalizable, and privacy-aware PPG authentication framework suitable for real-time mobile deployment remains an open challenge.

In addition to algorithmic design, PPG signal quality and acquisition parameters play a crucial role in authentication reliability. The PPG waveform is typically captured using optical sensors operating at different wavelengths offering varying sensitivity to skin tone, perfusion, and motion artifacts~\cite{allenPhotoplethysmographyOverview2007}. Multi-wavelength and multi-channel configurations, such as dual green or green–infrared pairs, have been shown to enhance robustness against ambient light interference and physiological noise~\cite{RayMultiWavelength2023}. Signal quality is also influenced by hardware parameters such as LED intensity, photodiode sensitivity, sampling rate, and frame rate, with typical acquisition frequencies ranging from 30 Hz for camera-based systems to over 125 Hz for dedicated PPG sensors~\cite{MejiaMejiaPPGSignalProcessing2021}. Proper calibration, filtering, and quality assessment algorithms (e.g., signal-to-noise ratio or template matching) are therefore critical to ensure consistent biometric performance under varying environmental and physiological conditions~\cite{liangOptimalFilterShort2018}.
However, most prior studies rely on broadband or color imaging setups, which remain sensitive to illumination variation, camera auto-exposure, and cross-device inconsistency. Studies rarely investigate monochrome illumination or narrow-band sensing, despite their potential to enhance signal-to-noise ratio and improve robustness in contactless or low-frame-rate scenarios.

To address the aforementioned challenges in PPG signal quality, motion artifacts, and limited generalization under low-frame-rate acquisition, we develop a compact preprocessing and learning framework tailored for monochrome blue-light PPG signals. The preprocessing pipeline incorporates baseline drift correction, PCA-based motion artifact suppression, adaptive bandpass filtering, resampling, and normalization to stabilize signals captured at 14 fps, ensuring consistent performance under realistic deployment conditions. Building on this, we propose a scalogram-based deep learning architecture that jointly models spatial, temporal, and frequency information. Integrated attention mechanisms refine the learned representations by emphasizing user-specific patterns, enhancing robustness to inter-subject variability and environmental noise. Overall, the proposed lightweight and cost-efficient authentication framework effectively bridges the gap between controlled experimental setups and real-world deployment, enabling stable, device-independent PPG-based authentication for next-generation mobile and embedded systems.

\subsection{Major Contributions}
This study advances PPG-based biometric authentication through a scalogram-driven hybrid deep learning framework optimized for monochrome blue-light acquisition. The proposed approach prioritizes computational efficiency, robustness to motion artifacts, and adaptability to real-world conditions. Leveraging attention-based mechanisms, the framework captures global dependencies and long-range temporal relationships while preserving fine-grained spatial and frequency information from scalogram representations. By jointly modeling spatial, temporal, and contextual features, the system achieves reliable and device-independent authentication suitable for mobile and embedded environments. Our major contributions are as follows:

\begin{enumerate}[label=\roman*)]
\item Developed a hybrid deep learning architecture combining CVT, ConvMixer, and LSTM modules to jointly learn spatial, contextual, and temporal representations.

\item Integrated attention-based refinement layers within the ConvMixer branch to emphasize user-specific physiological variations.

\item Benchmarked the new algorithm on a standard dataset (BIDMC) and showed that our system outperforms the state of the art. 

\item Achieved robust feature extraction and authentication from low-frame-rate monochrome PPG signals, demonstrating resilience under limited sampling conditions.
\end{enumerate}

The rest of this paper is organized as follows: Section II presents the detailed methodology, including data preprocessing, scalogram generation techniques, and the hybrid model architecture pipeline. Section III reports the experimental results and provides discussion, and concluding remarks are in Section IV.

\section{Methodology}
\begin{figure*}[htbp]
  \centering
  \includegraphics[width=\textwidth]{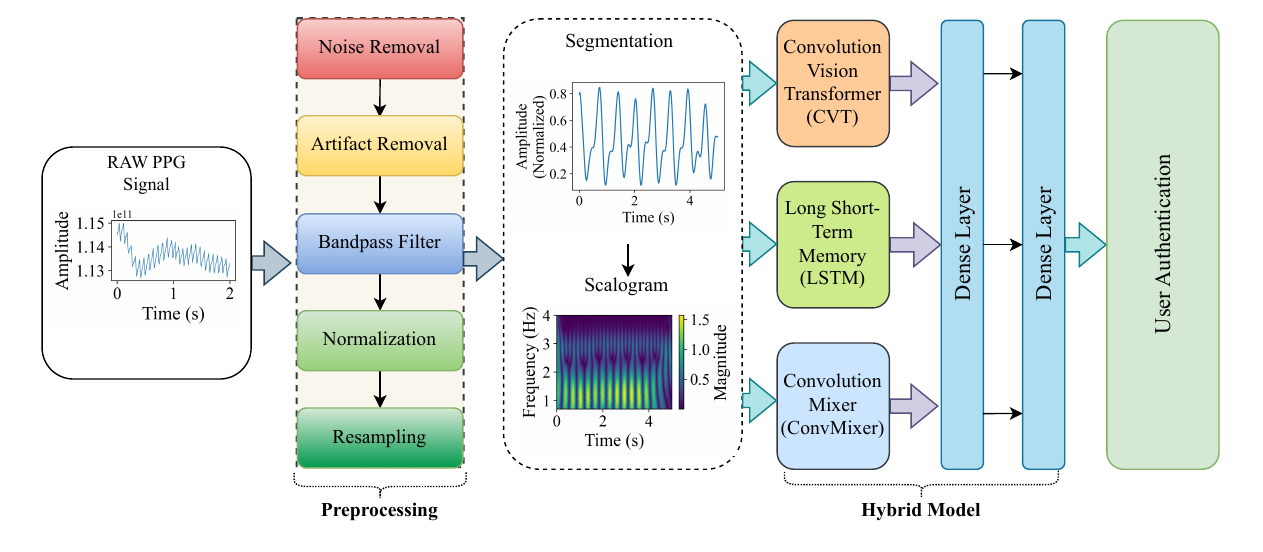}
  \caption{Flowchart of the proposed authentication system with the hybrid CVT–ConvMix–LSTM model.}
  \label{fig:flowchart}
\end{figure*}

This section outlines the end-to-end framework adopted for PPG-based biometric authentication. We first describe the data acquisition process, where fingertip videos are collected. Next, we detail the preprocessing pipeline designed to stabilize noisy signals, followed by the scalogram generation technique that converts one-dimensional PPG segments into time–frequency representations. Finally, we introduce a transformer-inspired hybrid framework designed to capture global context, local spatial details, and long-term temporal dependencies in PPG signals, thereby improving recognition accuracy under real-world conditions. The detailed framework is described in Fig \ref{fig:flowchart}.

\subsection{Data Acquisition}
\label{ssec:data_acquistion}

This study employs the Contactless Fingerprint Image Streams and Heart Rate (CFISHR) dataset \cite{Venkataswamy_CFISHR, OlugbenleAsilomar}. The dataset consists of high-resolution contactless fingerprint image streams of the right thumb from 46 participants, recorded at 14 frames per second using a monochrome blue-light scanner. The images captured have a resolution of 500 ppi. From these sequences, photoplethysmogram (PPG) signals were extracted by computing pixel-wise intensity variations across consecutive frames, thereby capturing blood volume changes over time. Data collection was conducted under controlled indoor conditions, with the scanner’s sealed chamber minimizing ambient light and background interference. Ground truth data is provided for each recording. 

\subsection{Preprocessing Raw PPG Signal}
\label{ssec:preprocessing}

\begin{figure}[!t]
    \centering
    \includegraphics[width=\columnwidth]{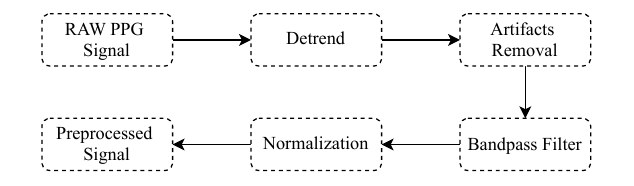}
    \caption{Preprocessing steps of raw PPG signal}
    \label{fig:preprocessing}
\end{figure}

Preprocessing is essential for enhancing the quality and consistency of photoplethysmography (PPG) signals prior to biometric analysis. The raw PPG signal, acquired at 14 Hz using a low-frame-rate imaging sensor, undergoes a structured preprocessing pipeline (Fig. \ref{fig:preprocessing}) to enhance signal quality and consistency.

Initially, detrending is applied to eliminate slow-varying baseline drifts caused by motion or contact pressure changes. Motion artifacts are then suppressed using dynamic thresholding, which clips extreme outliers. To isolate the frequency components associated with cardiovascular activity, a bandpass filter is used to suppress both low-frequency baseline wander and high-frequency noise. An optional moving average filter is applied to smooth minor fluctuations. The signal is then normalized using min-max scaling to ensure amplitude consistency across subjects. 

\subsection{Segmentation and Scalogram Generation}
\label{ssec:segmentation}
To facilitate time-frequency analysis, the continuous PPG signal, $x(t)$, is first divided into shorter overlapping frames to ensure local stationarity and robustness in feature extraction. Each signal is segmented into windows of duration $T_w$, with $\alpha$ overlap ratio. The stride size, $T_h$, defines how much the window moves forward after each step and can be expressed as:
\begin{equation}
    T_h = (1 - \alpha) T_w, 
\label{eq:stride}
\end{equation}
Therefore, the $k$-th segment can be expressed as
\begin{equation}
    x_k(t) = x(t)  w(t-kT_h),
\label{eq:segment}
\end{equation}
where $w(t)$ is a windowing function (e.g., rectangular or Hamming). Zero-padding is applied when the final segment does not completely fit within $T_w$ to ensure uniform segment lengths across all samples. This overlapping segmentation increases the number of training examples while preserving temporal continuity in the PPG signal \cite{STEPHANE200989}.
Each segment is then transformed into a two-dimensional scalogram using the Continuous Wavelet Transform (CWT). The CWT of a segmented signal $x_k(t)$ is defined as
\begin{equation}
W_x(s, \tau) = \frac{1}{\sqrt{s}} \int_{-\infty}^{\infty} x_k(t)\, \psi^*\left( \frac{t - \tau}{s} \right) dt, 
\label{eq:cwt}
\end{equation}
where $W_x(s, \tau)$ are the wavelet coefficients at scale $s$ and time-shift $\tau$, $\psi(t)$ is the wavelet function (e.g., Morlet), and $\psi^* (\cdot)$ denotes the complex conjugate of the wavelet.

Unlike the Short-Time Fourier Transform (STFT), CWT offers superior resolution for non-stationary physiological signals by adaptively adjusting the scale and position of the wavelet. This allows effective capture of both low- and high-frequency components. In this study, the Morlet wavelet is used due to its balanced time–frequency localization. After computing the wavelet coefficients, their absolute magnitude is taken to emphasize frequency content, and the resulting output is resized into a consistent 2D image format. These scalograms serve as rich visual representations of the temporal frequency dynamics in each segment and are used as input to our proposed authentication system. While typical heart rate frequencies range between 0.7 and 1.8 Hz \cite{ELHAJJ2020101870}, the scalograms span up to 4 Hz to capture higher-order harmonics and transient features, enhancing model robustness and individual discrimination in biometric authentication. 

\subsection{Hybrid Model Architecture: CVT–ConvMixer–LSTM}

We introduce a hybrid deep learning architecture that combines the strengths of Convolutional Vision Transformer (CVT), ConvMixer, and Long Short-Term Memory (LSTM) networks to perform robust biometric authentication using photoplethysmography (PPG) signals. The model exploits complementary spatial, frequency, and temporal feature representations for improved classification accuracy.

Each input consists of a segmented PPG signal, $\mathbf{x} \in \mathbb{R}^{T}$, where $T$ denotes the time length of the segment (e.g., 5 seconds). A time–frequency representation (scalogram) is generated from each segment using the Continuous Wavelet Transform (CWT) (see (\ref{eq:cwt})), resulting in a 2D image, $\mathbf{X}_{\text{scal}} \in \mathbb{R}^{H \times W}$. The 1D signal, $\mathbf{x}$, is retained as an input to the LSTM branch.

\subsubsection{CVT Branch}
The Convolutional Vision Transformer (CVT) component is designed to capture high-level spatial relationships from time–frequency representations (scalograms) of PPG signals. It extends the original Vision Transformer (ViT) by integrating convolutional operations into the token embedding and projection stages, allowing for localized spatial context modeling while maintaining global attention capabilities \cite{wu2021cvtintroducingconvolutionsvision}. The CVT combines convolutional layers with transformer-based attention to effectively encode local patterns and global context, enhancing the discriminative power of the model.

The process begins with a convolutional projection applied to the input scalogram $\mathbf{X}_{\text{scal}} \in \mathbb{R}^{H \times W}$:

\begin{align}
\mathbf{F}_{\text{CVT}} &= \text{Conv2D}(\mathbf{X}_{\text{scal}};k_c,s_c,p_c), \label{eq:conv2d} \\
\mathbf{E}_{\text{CVT}} &= \text{PosEmbed}(\mathbf{F}_{\text{CVT}}), \label{eq:posembed} \\
\mathbf{A}_{\text{CVT}} &= \text{MultiHeadAttention}(\mathbf{E}_{\text{CVT};}d_c,h_c), \text{ and } \label{eq:attention} \\
\mathbf{Z}_{\text{CVT}} &= \text{GlobalAvgPool}(\mathbf{A}_{\text{CVT}}), \label{eq:gap}
\end{align}
where $k_c$, $s_c$, $p_c$, $d_c$, and $h_c$ denote the kernel size, stride, padding, embedding dimension, and number of attention heads respectively for the CVT model. The parameters are empirically chosen based on model capacity and input resolution. However, they can be fine-tuned through hyperparameter optimization to balance accuracy and computational cost. The layer-wise dimension can be defined as
 $\mathbf{F}_{\text{CVT}} \in \mathbb{R}^{H_f\times W_f \times d_c}$, $\mathbf{E}_{\text{CVT}}  \in \mathbb{R}^{(H_fW_f) \times d_c}$,  $\mathbf{A}_{\text{CVT}}  \in \mathbb{R}^{(H_fW_f) \times d_c}$, and  $\mathbf{Z}_{\text{CVT}}  \in \mathbb{R}^{1 \times d_c}$.

The convolutional layer in (\ref{eq:conv2d}) represents the convolutional feature map obtained after downsampling the 256×256 input through two Conv2D layers with kernel size of 7 and stride value of 8, which extracts localized spatial features ($\mathbf{F}_{\text{CVT}} \in \mathbb{R}^{32\times 32 \times 64}$), capturing low-level textures and edge patterns essential to physiological signal representation. These feature maps are then enhanced with positional embeddings ($\mathbf{E}_{\text{CVT}} \in \mathbb{R}^{32\times 32 \times 64}$), as shown in (\ref{eq:posembed}) to retain spatial context, which is crucial for downstream attention operations. Multi-head self-attention as seen in (\ref{eq:attention}) is subsequently applied ($\mathbf{A}_{\text{CVT}}  \in \mathbb{R}^{1024 \times 64}$) to model global dependencies across spatial locations, leveraging multiple attention heads to capture diverse contextual relationships. Finally, global average pooling ($\mathbf{Z}_{\text{CVT}}  \in \mathbb{R}^{1 \times 64}$) in (\ref{eq:gap}) is applied to aggregate the features obtained after the application of the attention heads into a compact representation. The output, $\mathbf{Z}_{\text{CVT}}$, serves as the CVT branch feature vector, containing rich spatial relationships from the input scalogram.

\subsubsection{ConvMixer Branch}
The ConvMixer branch is designed to extract fine-grained spatial features from the input scalogram using patch-based convolutional processing. It combines depthwise and pointwise convolutions in a residual block structure to efficiently model spatial dependencies with minimal computational overhead \cite{Ibrahim_Sensors}.

The ConvMixer branch begins with a Conv2D projection that captures localized spatial features from the input scalogram:
\begin{align}
\mathbf{Z}_0 = \text{BN}(\sigma(\text{Conv2D}(\mathbf{X}_{\text{scal}};k_m,s_m,p_m))), 
\label{eq:convmixc}
\end{align}
where $k_m$, $s_m$, $p_m$, $h_m$, $\sigma (\cdot)$, and $BN$ denote the kernel size, stride, padding, number of attention heads, the activation function, and batch normalization, repsectively. The ConvMixer employs the Gaussian Error Linear Unit (GELU), a smooth and differentiable activation function known for its adaptive gating behavior, enabling effective handling of both positive and negative inputs and improving feature expressiveness in deep neural networks.

These features are passed through a series of ConvMixer blocks designed to extract fine-grained spatial relationships via patch-based processing. For each block, $l = 1 \dots L$, the following operations are performed:

\begin{align}
\mathbf{Z}_l &= \text{BN}\left(\sigma\left(\text{ConvDepthwise}\left(\mathbf{Z}_{l-1}\right)\right) + \mathbf{Z}_{l-1}\right), 
\label{eq:depthwise} \\
\mathbf{Z}_{l+1} &= \text{BN}\left(\sigma\left(\text{ConvPointwise}\left(\mathbf{Z}_l\right)\right)\right),
\label{eq:pointwise} \\
\mathbf{E}_{\text{CM}} &= \text{PosEmbed}\left(\mathbf{Z}_{l+1}\right),
\label{eq:posembed_cm} \\
\mathbf{A}_{\text{CM}} &= \text{MultiHeadAttention}\left(\mathbf{E}_{\text{CM}};d_m,h_m\right),
\label{eq:attention_cm} \\
\mathbf{Z}_{\text{CM}} &= \text{GlobalAvgPool}\left(\mathbf{A}_{\text{CM}}\right),
\label{eq:gap_cm}
\end{align}
where $d_m$ and $h_m$ denote the embedding dimension and number of attention heads respectively for the ConvMixer model.

Each block includes a depthwise convolution with residual connection (\ref{eq:depthwise}) followed by a pointwise convolution (\ref{eq:pointwise}); both normalized and activated. This structure allows the model to learn hierarchical patterns efficiently, while maintaining computational simplicity. The final feature maps are enriched with positional embeddings (\ref{eq:posembed_cm}) and passed through multi-head self-attention (\ref{eq:attention_cm}) to capture long-range spatial dependencies. The outputs are then aggregated via global average pooling (\ref{eq:gap_cm}) to form the ConvMixer feature vector $\mathbf{Z}_{\text{CM}} \in \mathbb{R}^{1 \times 32}$. 

\subsubsection{LSTM Branch}
LSTMs can learn long-term dependencies in sequential data by solving the vanishing gradient problem. This makes them effective for capturing temporal patterns which could be used in different tasks \cite{maTSLSTMTemporalInceptionExploiting2017,
degeestModelingTemporalStructure2018}. 

To capture temporal dependencies in PPG-based biometric authentication, the segmented signal $x_k (t)$ is processed by an LSTM layer that operates over time to extract robust temporal features. At time step $t$, the network receives the current input $x_k{(t)}$ along with the hidden state $h^{(t-1)}$ and cell state $c^{(t-1)}$ from the previous step. It then computes the forget gate $f^{(t)}$, input gate $i^{(t)}$, output gate $o^{(t)}$, candidate cell state $\tilde{c}^t$, updated cell state $c^{(t)}$, and new hidden state $h^{(t)}$, enabling effective learning of long-range dependencies in sequential physiological data. The update rules \cite{weiSelfAttentionBiLSTMNetworks2021} are defined as:
\begin{equation}
\left\{
\begin{aligned}
f^{(t)} &= \sigma_s\left(\mathbf{W_f}[h^{(t-1)}, x_k{(t)}] + b_f\right), \\
i^{(t)} &= \sigma_s\left(\mathbf{W_i}[h^{(t-1)}, x_k{(t)}] + b_i\right), \\
o^{(t)} &= \sigma_s\left(\mathbf{W_o}[h^{(t-1)}, x_k{(t)}] + b_o\right), \\
\tilde{c}^{(t)} &= \tanh\left(\mathbf{W_c}[h^{(t-1)}, x_k{(t)}] + b_c\right), \\
c^{(t)} &= i^{(t)} \odot \tilde{c}^{(t)} + f^{(t)} \odot c^{(t-1)}, \text{ and } \\
h^{(t)} &= o^{(t)} \odot \tanh\left(c^{(t)}\right),
\end{aligned}
\right.
\label{eq:lstm}
\end{equation}
where $\sigma_s$ is the activation function, $\odot$ is the element-wise (Hadamard) product, and $W$  and $b$  are the learnable parameters. After the final time step, the last hidden state ($\mathbf{H}_{\text{final}} = h^{(t)}$) serves as the temporal feature vector $Z_{\mathrm{LSTM}}$ for fusion with spatial features from the CVT and ConvMixer branches. 

\subsubsection{Feature Fusion and Authentication}
Following the extraction of spatial and temporal feature embeddings from the CVT, ConvMixer, and LSTM branches, a unified fusion strategy is employed to generate a comprehensive biometric representation. The outputs of the three branches are concatenated along the feature dimension to form the fused representation:
\begin{equation}
\mathbf{F}_{\text{concat}} = \text{concat}(\mathbf{Z}_{\text{CVT}}, \mathbf{Z}_{\text{CM}}, \mathbf{Z}_{\text{LSTM}}),\label{eq:feature_concat}
\end{equation}
where the concat$(\cdot)$ operation concatenates the vectors in a row-wise fashion, effectively merging the spatial and temporal feature representations from the three branches into a unified feature vector.

The fused feature vector is then passed through a fully connected dense layer followed by a softmax activation function to yield normalized class probabilities:
\begin{equation}
\hat{y} = \text{softmax}(\text{Dense}(\mathbf{F}_{\text{concat}})), \label{eq:soft_max}
\end{equation}
where $\hat{y} \in \mathbb{R}^C$ represents the predicted probability distribution over $C$ enrolled subjects. The softmax layer maps the fused features to a normalized probability distribution, enabling final user authentication by identifying the subject with the highest confidence score. 

\subsubsection{Enrollment and Authentication}
\label{sssec:enrollment_authentication}
During the enrollment phase, a subject’s PPG signal segments are first transformed into their corresponding fused feature vectors $\mathbf{F}_{\text{concat}}$, as defined in (\ref{eq:feature_concat}). For each subject $i$, an enrolled template vector $\mathbf{T}_i$ is computed by averaging the embeddings of multiple training segments:
\begin{equation}
\mathbf{T}_i = \frac{1}{N_i} \sum_{n=1}^{N_i} \mathbf{F}_{\text{concat},i}^{(n)},
\label{eq:template}
\end{equation}
where $N_i$ is the number of enrolled samples for subject $i$, and $\mathbf{F}_{\text{concat},i}^{(n)}$ is the fused embedding obtained from the $n^{th}$ enrolled segment of that subject. This template serves as the user’s biometric signature in the feature space.
During the authentication phase, a test segment is processed through the trained model to obtain its fused embedding $\mathbf{F}_{\text{test}}$. Authentication is then performed by computing the cosine similarity between $\mathbf{F}_{\text{test}}$ and each enrolled template $\mathbf{T}_i$:
\begin{equation}
S_i = \frac{\mathbf{F}_{\text{test}} \cdot \mathbf{T}_i}{\|\mathbf{F}_{\text{test}}\| \, \|\mathbf{T}_i\|},
\label{eq:similarity}
\end{equation}
where $S_i$ represents the similarity score for subject $i$. The test sample is estimated as belonging to subject $\hat{i}$ if:
\begin{equation}
\hat{i} = \argmax_i S_i, \quad \text{subject to } S_i \geq \tau,
\label{eq:decision}
\end{equation}
where $\tau$ is a predefined similarity threshold determined empirically (e.g., at equal error rate). Note that if $\hat{i} = i$, authentication is successful.  
\section{Result and Discussion}
This section presents the experimental results of the proposed framework, outlining each stage sequentially—from signal preprocessing and scalogram generation to model training and authentication.
\subsection{Preprocessed Signals}
\begin{figure*}[htbp]
  \centering
  \includegraphics[width=\textwidth]{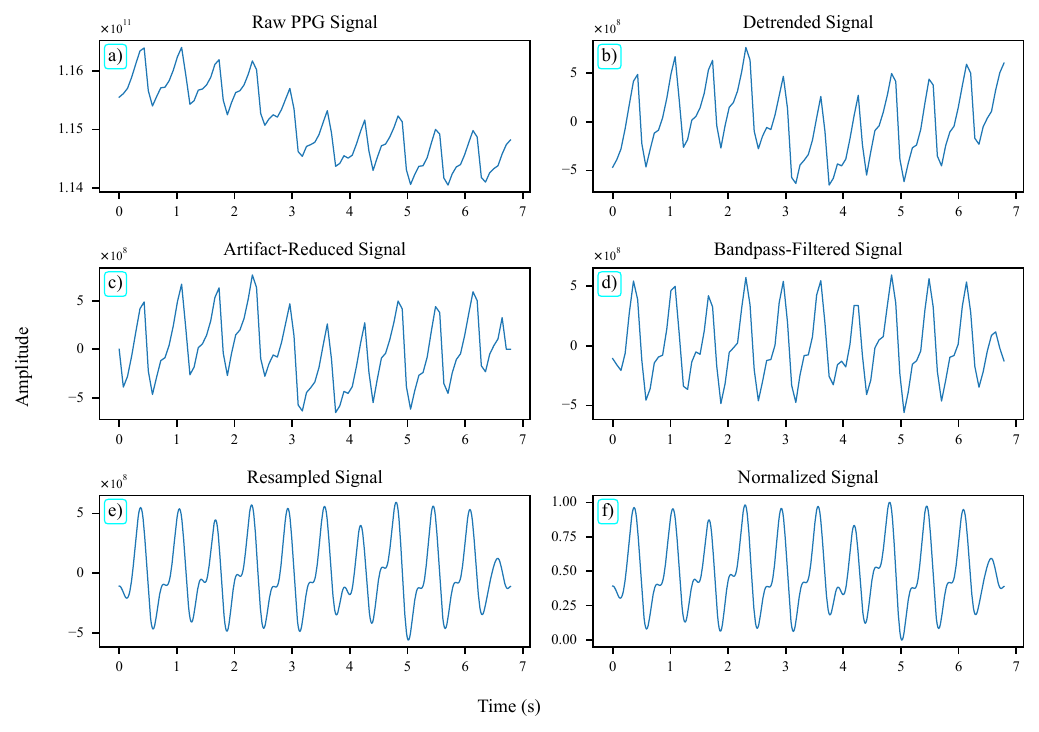}
  \caption{Preprocessing pipeline for PPG signals: (a) raw PPG signal, (b) detrended signal, (c) artifact-reduced signal, (d) bandpass-filtered signal, (e) resampled signal, and (f) normalized signal. Each stage progressively enhances signal quality and prepares the data for feature extraction.}
  \label{fig:preprocessed_resultv2}
\end{figure*}
The preprocessing pipeline progressively enhances the raw PPG signal by removing noise and preparing it for robust feature extraction. Each participant in our experiment, recorded approximately 70 seconds of continuous right thumb fingerprint video, producing a continuous sequence of frames from which PPG signals were extracted. The extracted PPG signals were sampled at $f_s = 14,\text{Hz}$, yielding 980 samples per recording. First, the raw signal (Fig. \ref{fig:preprocessed_resultv2}a) is detrended to eliminate slow-varying baseline drifts caused by motion or illumination changes (Fig. \ref{fig:preprocessed_resultv2}b). Next, extreme outliers are clipped during artifact removal (Fig. \ref{fig:preprocessed_resultv2}c), reducing the influence of abrupt disturbances. A bandpass filter (Fig. \ref{fig:preprocessed_resultv2}d) is then applied to isolate the frequency band of interest (0.7–4Hz), corresponding to physiological heart rate components, while suppressing low- and high-frequency noise. 
To improve temporal resolution and ensure sufficient sampling density for time–frequency analysis, the filtered PPG signals are resampled (Fig. \ref{fig:preprocessed_resultv2}e) using Fourier-based interpolation \cite{Sacchi1998}. The sampling rate is increased from the original $f_s = 14,\text{Hz}$ to $f_r = 5f_s = 70,\text{Hz}$, resulting in $L = 4900$ samples per subject. This upsampling factor preserves signal morphology while providing finer temporal detail required for continuous wavelet transform (CWT)–based scalogram generation.
Finally, amplitude normalization (Fig. \ref{fig:preprocessed_resultv2}f) scales the signal to a uniform range, ensuring consistency across subjects and enabling reliable inputs to the hybrid deep learning model. 
\subsection{Scalogram Generation from Segmented Signals}
\begin{figure}[htbp]
    \centering
    \includegraphics[width=\columnwidth]{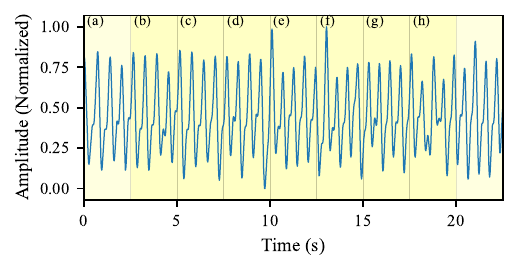}
    \caption{Segmentation of the PPG signal into fixed-length windows with 50\% overlap. Each segment spans 5 seconds at the resampled frequency of 70 Hz, resulting in 350 data points per window.}
    \label{fig:segmentation}
\end{figure}
\begin{figure}[htbp]
    \centering
    \includegraphics[width=\columnwidth]{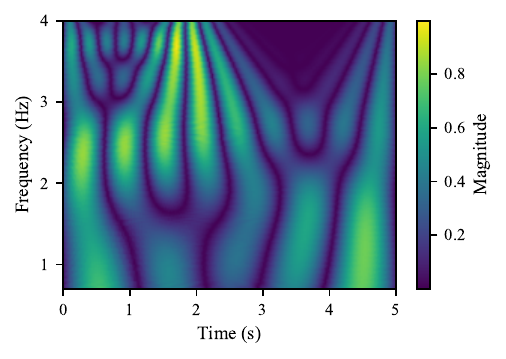}
    \caption{Scalogram of a segmeneted PPG signal using continuous wavelet transform (CWT).}
    \label{fig:scalogram}
\end{figure}

To enable time–frequency analysis, each preprocessed PPG signal is segmented into fixed-length windows of $T_w= 5s$ ($N_w = 350$ samples) with an overlap of $\alpha= 50\%$, and a hop size of $T_h = 2.5\,\text{s}$ ($N_h = 175$ samples). The number of segments can be computed by 
\( N = \left\lfloor \frac{L - N_w}{N_h} \right\rfloor + 1 \), 
yielding \( N = 27 \) segments per subject.The use of overlapping windows ensures continuity between adjacent segments and enhances the robustness of feature extraction. Segment lengths are standardized using zero-padding where necessary to maintain uniformity across samples. As an example, illustrated in Fig.~\ref{fig:segmentation}, a PPG signal of 23\,s duration is partitioned using the defined parameters ($T_w= 5s$ , $\alpha= 50\%$), resulting in seven overlapping segments suitable for subsequent wavelet-based analysis. For a given segment, $x_k(t)$, the CWT coefficients, $W_x(s,\tau)$, capture how the energy of the signal is distributed across scales, $s$, (inversely related to frequency) and temporal positions, $\tau$, as described in (\ref{eq:cwt}). Taking the absolute magnitude of these coefficients produces the scalogram, a two-dimensional representation encoding the spectral evolution of the signal over time. This representation preserves transient features and harmonics beyond the fundamental heart-rate frequency, providing a rich input for deep feature extraction. 

As illustrated in Fig.~\ref{fig:scalogram}, the scalogram of a 5-second PPG segment reveals dominant frequency components concentrated below $2\,\text{Hz}$, consistent with the physiological heart rate range. Higher-order harmonics and transient variations are also visible, which are difficult to capture in the time domain. Converting the PPG into scalograms enables the application of convolution-based architectures, which are highly effective in extracting discriminative features from image-like representations. This transformation bridges the gap between physiological signal processing and modern computer vision methods, enhancing robustness and accuracy in biometric authentication.

\subsection{Model Training}
We use the CFISHR dataset \cite{Venkataswamy_CFISHR} to validate the proposed hybrid biometric model. The CFISHR dataset consists of contactless fingerprint image streams and synchronized PPG signals captured from 46 participants aged 18–33 years. Each recording was captured using a monochrome camera at 14 fps, producing grayscale intensity sequences from which PPG waveforms were extracted. The resulting PPG signals have a sampling rate of $f_s = 14$ Hz and serve as the input for subsequent preprocessing and model training.
To evaluate authentication performance, 80\% of the segmented scalograms from each subject were assigned for training and the remaining 20\% for testing, ensuring balanced subject representation across both sets. Feature extraction was carried out using the proposed hybrid architecture (Section~II-D), which integrates a Convolutional Vision Transformer (CVT) branch, a ConvMixer branch, and a bidirectional LSTM branch. The CVT branch employs a convolutional projection into a $64$-dimensional embedding, followed by tokenization into $16$ patches and refinement through multi-head self-attention with four heads, enabling the capture of both local and global spatial dependencies. The ConvMixer branch leverages depthwise separable convolutions with $32$ filters and kernel size of $5$ to extract localized spatial features, further enhanced by normalization and attention. In parallel, the LSTM branch incorporates $64$ hidden units to model long-term temporal dependencies, effectively preserving physiological rhythm information. Outputs from all three branches are concatenated and projected into a unified $64$-dimensional embedding, forming robust subject-specific biometric templates that fuse spatial, spectral, and temporal information for authentication.  

\subsection{Authentication Results}
To evaluate the performance of the proposed PPG-based authentication system, we report several widely used biometric performance metrics, including Accuracy, Sensitivity (Recall), Specificity, Equal Error Rate (EER), and Area Under the ROC Curve (AUC). These measures provide a comprehensive assessment of the system’s ability to correctly authenticate genuine users while rejecting impostors. These metrics are defined as:

\begin{align}
\text{Accuracy} &= \frac{TP + TN}{TP + TN + FP + FN}, \label{eq:accuracy} \\
\text{Sensitivity (Recall)} &= \frac{TP}{TP + FN}, \label{eq:sensitivity} \\
\text{Specificity} &= \frac{TN}{TN + FP}, \label{eq:specificity}
\end{align}

\noindent
where $TP$, $TN$, $FP$, and $FN$ denote the number of true positives, true negatives, false positives, and false negatives, respectively.
%
The Area Under the ROC Curve (AUC) is defined as the integral of the True Positive Rate (TPR) with respect to the False Positive Rate (FPR), which summarizes the overall discriminative capability of the model across all thresholds:
\begin{equation}
\text{AUC} = \int_{0}^{1} TPR(FPR)\, d(FPR). 
\end{equation}

\begin{table}[t]
\centering
\caption{State-of-the-art biometric authentication system compared to the CVT-ConvMixer-LSTM model on BIDMC\cite{BIDMC} and  Heart Rate (CFISHR) \cite{Venkataswamy_CFISHR} datasets}
\resizebox{\columnwidth}{!}{%
\begin{tabular}{lccccc}
\hline
\textbf{Studies} & \textbf{Accuracy} & \textbf{Sensitivity} & \textbf{Specificity}  & \textbf{AUC} \\ \hline
CorNET \cite{BiswasCorNET}        & 0.96 & 0.93 & 0.92 & 0.90 \\
CNN-LSTM \cite{Hwang10.1109}       & 0.96 & 0.93 & 0.95  & 0.93 \\
Fuzzy-Min-Max-NN \cite{YuzeZhangSPIE} & 0.78 & 0.80 & 0.82 & 0.81 \\
CVT-ConvMixer \cite{Ibrahim_Sensors}              & 0.95 & 0.97 & 0.95  & 0.96 \\ 
CVT-ConvMixer-LSTM (BIDMC)
\cite{BIDMC}      & 0.98 & 0.96 & 0.97  & 0.98 \\
\hline
CVT-ConvMixer-LSTM 
(CFISHR)
\cite{Venkataswamy_CFISHR}.        & 0.98 & 0.95 & 0.93  & 0.95 \\ \hline
\end{tabular}%
}
\label{tab:soa_biometric}
\end{table}

The proposed hybrid architecture couples time–frequency scalogram–based spatial learning with explicit temporal modeling to exploit the full structure of the biometric signal. To assess the contribution of each component, we compared three configurations: LSTM-only, CVT+ConvMix, and the proposed hybrid model. Using the CFISHR dataset, the LSTM configuration achieved an authentication accuracy of 89.48\%, while CVT+ConvMix reached 95.56\%. The combined hybrid architecture outperformed both, achieving an authentication accuracy of 97.68\%, thus confirming the complementary strengths of spatial and temporal modeling. Other evaluation metrics, including sensitivity, specificity, and AUC as shown in Table~\ref{tab:soa_biometric}, exhibit consistent improvement, further validating the hybrid design.

To validate the effectiveness of our proposed model, we compared its performance against 
state-of-the-art biometric identification approaches reported in recent literature. 
Table~\ref{tab:soa_biometric} summarizes accuracy, sensitivity, specificity, and AUC across different deep learning models. 
We note here that these models have been evaluated on the BIDMC dataset \cite{BIDMC}, and the performance of our algorithm, in the final two rows, was obtained using separate processing pipelines on the BIDMC and CFISHR datasets. 

\begin{figure}[t]
  \centering
   \includegraphics[width=\columnwidth]{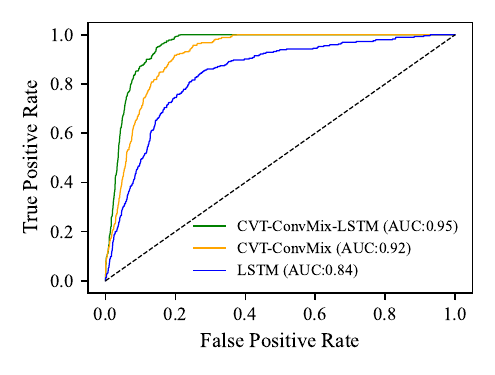}
  \caption{ROC curve of the proposed hybrid model.}
  \label{fig:roc}
\end{figure}

The Receiver Operating Characteristic (ROC) analysis further substantiates the discriminative capability of the proposed architecture in distinguishing between genuine and impostor attempts. As depicted in Fig.~\ref{fig:roc}, the baseline LSTM yields an AUC of 0.84, reflecting moderate separability but limited robustness in capturing complex scalogram patterns. Incorporating transformer-based spatial learning through CvT-ConvMix elevates the AUC to 0.92, highlighting the benefit of combining global attention with localized convolutional representations. The full CvT-ConvMix-LSTM further enhances performance, achieving an AUC of 0.95. Its ROC curve exhibits a steep ascent near the origin, indicating the ability to detect the majority of genuine attempts at very low false positive rates. This property is particularly critical for authentication scenarios, where minimizing false acceptances is essential. Collectively, the ROC results align with the accuracy improvements reported in Table~\ref{tab:soa_biometric}, reinforcing that joint spatial–temporal reasoning provides superior robustness and state-of-the-art separability across varying decision thresholds.
These results show that the proposed CVT-ConvMixer-LSTM model consistently outperforms conventional deep learning approaches. In particular, it achieves the high accuracy and AUC on both BIDMC and CFISHR datasets, demonstrating strong generalization across distinct acquisition domains. The improvement highlights the effectiveness of hybrid spatial–temporal feature fusion in capturing discriminative cardiovascular patterns for reliable biometric authentication.

\section{Conclusions}
This study establishes a robust framework for PPG-based biometric authentication by fusing advanced deep learning architectures with time–frequency feature representations. The proposed hybrid model leverages the complementary strengths of the Convolutional Vision Transformer (CVT) for global attention, and ConvMixer for localized feature learning, which together enhance representational capacity, robustness, and include an LSTM branch, which effectively captures temporal correlations that are lost during convolutional operations. By preserving sequential dynamics in the PPG signal, the LSTM-enhanced hybrid model achieves high authentication accuracy, underscoring the critical role of temporal dependency modeling in biometric feature extraction.
The promising results validate the practicality of our approach for real-world identity verification, where accuracy and resilience are paramount. 

Future work includes the exploration of the fusion of PPG with additional physiological modalities such as ECG or accelerometer data, creating more comprehensive multimodal authentication systems. Moreover, investigating lightweight and energy-efficient variants of the hybrid model would support deployment in wearable and mobile platforms.

\bibliographystyle{IEEEtran}
\bibliography{bibtex/bib/references, bibtex/bib/IEEEabrv, bibtex/bib/IEEEexample}

%
\begin{IEEEbiography}[{\includegraphics[width=1in,height=1.25in,clip,keepaspectratio]{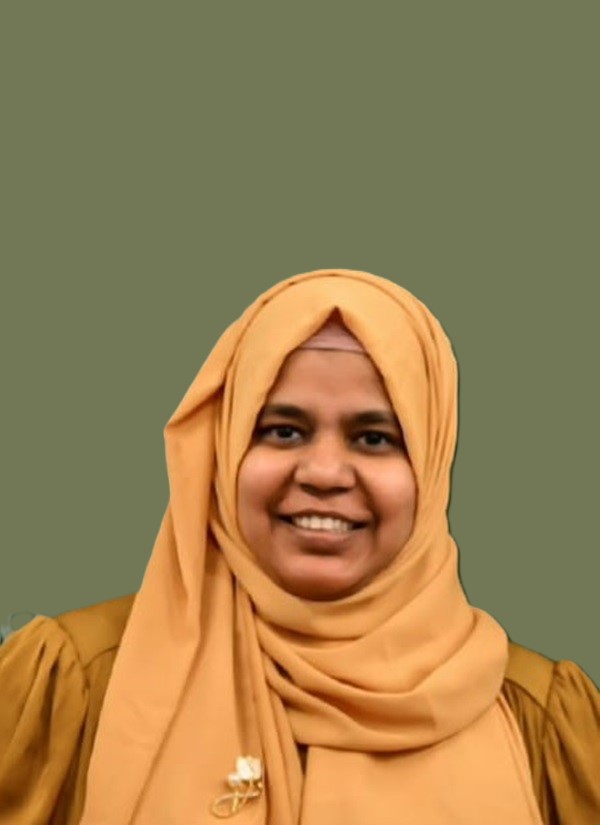}}]{Arfina Rahman}
received the B.Sc. degree from Dhaka International University, Dhaka, Bangladesh. She is currently pursuing the graduate degree with the Department of Electrical and Computer Engineering, Clarkson University, Potsdam, NY, USA. Her research interests include the design and development of algorithms for biometric authentication, biomedical engineering, signal processing and behavioral biometrics.
\end{IEEEbiography}

\begin{IEEEbiography}[{\includegraphics[width=1in,height=1.25in,clip,keepaspectratio]{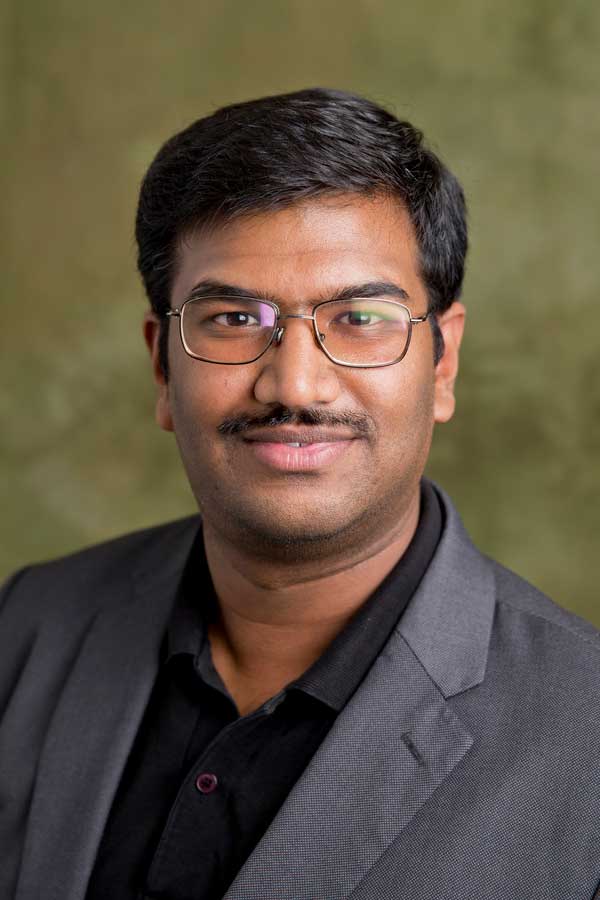}}]{Mahesh Banavar}
received the B.E. degree in
telecommunications engineering from Visvesvaraya
Technological University, in 2005, and the M.S. and
Ph.D. degrees in electrical engineering from Arizona
State University, in 2007 and 2010, respectively. He
is currently a Professor with the Department of ECE, Clarkson University, Potsdam, NY,
USA. His interests include node localization, detection and estimation algorithms, and user-behaviorbased cybersecurity applications.
\end{IEEEbiography}









\end{document}